\documentclass[%
 preprint,
superscriptaddress,
 amsmath,amssymb,
 aps,
]{revtex4-1}

\usepackage{graphicx}
\usepackage{dcolumn}
\usepackage{bm}
\usepackage{hyperref}

\usepackage{tipa} 

\usepackage{longtable} 
\usepackage{multirow}
\usepackage[table]{xcolor}

\usepackage{rotating}
\usepackage{url}
\usepackage[table]{xcolor}
\usepackage{multirow}
\usepackage{epigraph}
\usepackage{hyperref}

\begin{document}

\author{Marius Mavridis}
\affiliation{%
Institute for Cross-Disciplinary Physics and Complex Systems IFISC (UIB-CSIC), Campus Universitat de les Illes Balears, E-07122 Palma de Mallorca, Spain
}
\affiliation{Université Paris-Saclay, ENS Paris-Saclay, DER de Physique, 91190, Gif-sur-Yvette, France}
\author{Juan De Gregorio}

\author{Ra\'ul Toral}

\author{David S\'anchez}

\affiliation{%
Institute for Cross-Disciplinary Physics and Complex Systems IFISC (UIB-CSIC), Campus Universitat de les Illes Balears, E-07122 Palma de Mallorca, Spain
}

\title{Phonological distances for linguistic typology and the origin of Indo-European languages}


\date{\today}

\begin{abstract}
We show that short-range phoneme dependencies encode large-scale patterns of linguistic relatedness, with direct implications for quantitative typology and evolutionary linguistics. Specifically, using an information-theoretic framework, we argue that phoneme sequences modeled as second-order Markov chains essentially capture the statistical correlations of a phonological system. This finding enables us to quantify distances among 67 modern languages from a multilingual parallel corpus employing a distance metric that incorporates articulatory features of phonemes. The resulting phonological distance matrix recovers major language families and reveals signatures of contact-induced convergence. Remarkably, we obtain a clear correlation with geographic distance, allowing us to constrain a plausible homeland region for the Indo-European family, consistent with the Steppe hypothesis.
\end{abstract}

\maketitle

\section{\label{sec:intro}Introduction}

The computation of language distances offers a useful approach that permits to quantitatively explore language relationships based on spelling, phonetic, syntactic or lexical similarities. The topic flourished in the field of dialectometry, which seeks to quantify the variability of dialects within a geographical area~\cite{goebl2006recent}. Thus, distances are evaluated between feature vectors representing phonetic symbols using metrics such as Levenshtein, Manhattan or Euclidean measures~\cite{nerbonne1997measuring}. This method can be employed to make a direct comparison between dialects~\cite{nerbonne2001computational}, even for noncontiguous varieties that arise as a result of migration.

An interesting application lies within language acquisition and learning. The distance from L1 (native tongue) to L2 (second language) impacts learner effort in terms of fluency, cognitive load, etc. This approach thus converts an abstract concept into a quantitative measure that can be used to design efficient teaching strategies. For instance, when L2 is English, Swedish learners attain proficiency much faster than Japanese learners~\cite{chiswick2005linguistic}. In multilingual societies, these distance measures can help policymakers better understand the complexity of linguistic adaptation to more effectively address issues faced by minority language communities~\cite{esser2006migration}.

Since the evaluation of linguistic distance can then be applied to assess language relatedness, this finds interesting applications in both language typology~\cite{levshina2022corpus} and historical linguistics~\cite{jenset2017quantitative}. On the one hand, quantitative metrics can measure how far apart two languages are, which leads to the generation of maps of similarities~\cite{gamallo2017language}. This is a data-driven (or, more precisely, corpus-driven) approach that can be benchmarked against the Swadesh list or related databases~\cite{brown2008automated}. Usage of improved metrics can effectively distinguish related languages from unrelated ones~\cite{wichmann2010evaluating}. On the other hand, the Levenshtein distance can be used to obtain language trees such as that of the Indo-European family~\cite{serva2008indo}. Then, tools from statistical mechanics and population dynamics can be employed to model language evolution~\cite{petroni2008language}. Further phylogenetic relations can be established from more complex, hypothesis-testing models. In this context, language distances have been proven useful even in low-resource languages~\cite{gamallo2020measuring,estarrona2023measuring}.

Here, we investigate the language clusters resulting from an analysis of interlinguistic phonological distances. This problem has been previously addressed with various methods~\cite{marian2012clearpond,eden2018measuring,lara2022multiplex}. We apply an information-theoretic approach that models sequences of phonemes as higher-order Markov chains. Further, our work is also motivated by the recent finding~\cite{de2024exploring} that there exists a correlation between morphosyntactic and geographic distances based on Universal Dependencies~\cite{de2021universal,li2025measuring}. As shown below, we also find such a correlation but in the phonological domain, which is important in the ongoing discussion on the dependence between language relatedness and spatial proximity~\cite{jeszenszky2017exploring,jager2018global}. When the correlation is examined between a genetically related group of languages, this dependence may help determine the homeland for their common ancestor~\cite{wichmann2010homelands}. We pursue this idea to infer from our results the origin of the Indo-European family, a question that has lately attracted much attention~\cite{bouckaert2012mapping,chang2015ancestry,heggarty2023language}.

\section{\label{sec:methods} Materials and Methods}

\subsection{Dataset}

We analyze the phonological variation across languages with a parallel corpus. This approach limits biases due to different genres or registers while ensuring at the same time a similar sample length. We select the Bible due to its accessibility for natural language processing and availability in many languages. Indeed, a Bible corpus has served as the essential raw material for calculating language distance of the lexicosemantic type~\cite{sanchez2023ordinal}. Here, we will focus on the phonological domain by means of a transcription scheme that we detail below.

Our corpus consists of 129~translations of the Bible, obtained from different sources~\cite{christodouloupoulos2015massively,biblecom}.
Despite the advantages outlined above, our corpus also has limitations. A variety of translation styles exists, depending on the epoch of translation and the dominant trend at that time. The oldest versions tend to adopt a more literal, or word-to-word approach, whereas the more modern ones usually focus on conveying meaning, to the detriment of original style. We thus select the oldest version available for each language. This introduces a bias towards a more archaic language, not necessarily entirely compatible with modern transcription rules. Additionally, as a religious book, the Bible is also biased on a lexical level. However, since our method is based on a statistical comparison of phoneme distributions, we do not expect that these limitations will significantly alter the results.

We generate the phonemic transcriptions for our Bible corpus using Phonemizer (the eSpeak backend~\cite{bernard_phonemizer_2021}) and Epitran~\cite{mortensen_epitran_nodate}, which transform text strings into the International
Phonetic Alphabet. Due to the language availability in these tools, the obtained transcriptions correspond to 67~languages (see Appendix~\ref{app_data} for the full list)~\cite{dataset}.
For definiteness, we remove suprasegmental
features such as tones, as well as diacritics, indicating subtle articulatory differences. We also disregard phonetic length, implying that both long and short vowels are considered the same sound whereas long consonants are analyzed as double. Nevertheless, we keep aspiration, palatalization, pharyngealization, murmur (for consonants) and nasalisation (for vowels), since these distinctions reflect the phonemic identity of certain languages.

We next validate the transcriptions using the WikiPron database~\cite{lee-etal-2020-massively}, which provides the phonological transcriptions (both broad and narrow) featured in Wiktionary~\cite{wiktionary}. For a given language, we compare words present in both our corpus and WikiPron.
The percentage of matches between Phonemizer and Wiktionary transcriptions ranges, for 21~languages tested, from 13\% (for Hindi and Gujarati) to 99\% (for Estonian and Finnish), with only 9~languages out of 21~reaching more than 50\%. We then compute the Levenshtein distance between nonmatches. This metric quantifies the divergence between transcriptions as the minimum number of single-phoneme operations (insertions, deletions or substitutions) required to transform one transcription into the other. We find in a majority of cases that this distance is
equal to~1. As a consequence, both transcriptions only differ by one operation. In most of these cases (88\% on average), the operation to be performed is a substitution, meaning that Phonemizer and Wiktionary disagree on the interpretation of one given sound. For each language, we identify that the main disagreements can be explained by one of these factors: (i) allophonic variation, i.e., two sounds are allophones of the same phoneme, thus making the two transcriptions valid, at least with a broad approach (for example, Phonemizer often transcribes the grapheme \textit{a} as /\textipa{a}/, while WikiPron gives /\textipa{\ae}/, in words such as \textit{gap} or \textit{sack}: these two phonemes are allophones in English Received Pronunciation); (ii) regional differences, i.e., if one considers each language as a macrolanguage encompassing features from its different varieties both transcriptions are acceptable (for instance, WikiPron gives /\textipa{k\textepsilon @}/ as a transcription for the word \textit{care} while Phonemizer transcribes it as /\textipa{ke@}/, which is found in New-Zealand English); and (iii) an actual error by Phonemizer (for example /\textipa{\textturnr a\texttheta}/ for \textit{wrath}, which is actually pronounced with /\textipa{\ae}/, /\textipa{6}/, /\textipa{A}/ or /\textipa{O}/ depending on the variety). This last case is subtler since each error is both phoneme- and language-dependent. Therefore, we modify the data as less as possible, only correcting Phonemizer transcriptions when the error is manifestly evident.

\subsection{Information-theoretic approach}
We treat our transcribed sequences of phonemes as a stochastic process whose states are the possible symbols in a language phonemic inventory of size $L$, where $L$ ranges between 26 (for Finnish) and 59 (for Hindi) in our dataset.
We want our method to be less sensitive to well known language-specific
segmentation issues with phonological words~\cite{dixon2003word}. To this end,
we omit word boundaries and the entire transcribed sequence becomes a single string of phonemes. This also has the advantage of capturing correlation across words, potentially providing further insight into the phonological structure of a given language.

We divide the string in overlapping blocks of size $r$, hereafter termed $r$-phones for short. We count the occurrence number for each $r$-phone, thus estimating their probability distribution for each language. Our estimates will be reliable provided that $r \leq r_{\rm max}$, with the maximum block size $r_{\rm max} \ll N$, $N$ being the string length. Below, we will give a more precise expression for $r_{\rm max}$.

We model the stochastic process as a discrete high-order Markov chain~\cite{raftery1985model}. A Markov process with states $x_1,...,x_L$ has order or memory $m$ if, for all $n \geq m$, the $n$-th order transition probabilities reduce to the $m$-th order:
 \begin{equation}
p(x_{i_{n+1}}|x_{i_1},...,x_{i_n}) = p(x_{i_{n+1}}|x_{i_{n-m+1}},...,x_{i_n})
 \end{equation}
The value $m = 0$ corresponds to the case where all states are independent, while $m=1$ is the classical Markovian case.

The statistical properties of a Markov process can be described with the block entropy
\begin{equation} \label{eq_Hr}
H_r = -\sum_{i=1}^{L^r}p(B_i)\ln(p(B_i))\,,
\end{equation}
where $B_i = x_{j_1}...x_{j_r}$ is one of the $L^r$ possible blocks (in our case, $r$-phones) of $r$ consecutive states
and $p(B_i)$ its probability.
The negative second discrete derivative of $H_r$,
\begin{equation} \
\mathcal{G}_u = -(H_{u+2} - 2H_{u+1} + H_u)\,,
\end{equation}
is the predictability gain~\cite{crutchfield_regularities_2003}, which measures the average gain in information when taking into account $u+1$ instead of $u$th-order transition probabilities, for integer $u\ge 0$.
One can hence show that a Markov process has memory $m$ if and only if $\mathcal{G}_u = 0$ for all $u\geq m$~\cite{de2025information}. 

The number of possible blocks of size $r$ is $L^r$ whereas the amount of available data becomes $N - r + 1$. Thus, as $r$ increases, the undersampled regime is eventually reached.
It follows that the block entropy $H_r$ can be reliably estimated for $r \leq r_{\rm max} = \lfloor{\frac{\ln(N)}{\ln(L)}}\rfloor$~\cite{de2022improved}. Since $N\simeq 10^6$ for the languages in our dataset, $r_{\rm max}$ is between 3 and 4. We compute $p(B_i)$ for block sizes from 1 to 5 and insert the obtained values in Eq.~\eqref{eq_Hr} with the convention $H_0=0$. We numerically calculate entropies using the Nemenman-Shafee-Bialek estimator~\cite{nemenman2001entropy}, which is one of the available estimators with least mean squared error for correlated data~\cite{de_gregorio_entropy_2024}. 

\begin{figure}[t]
    \centering
    \includegraphics[width=.85\textwidth,clip]{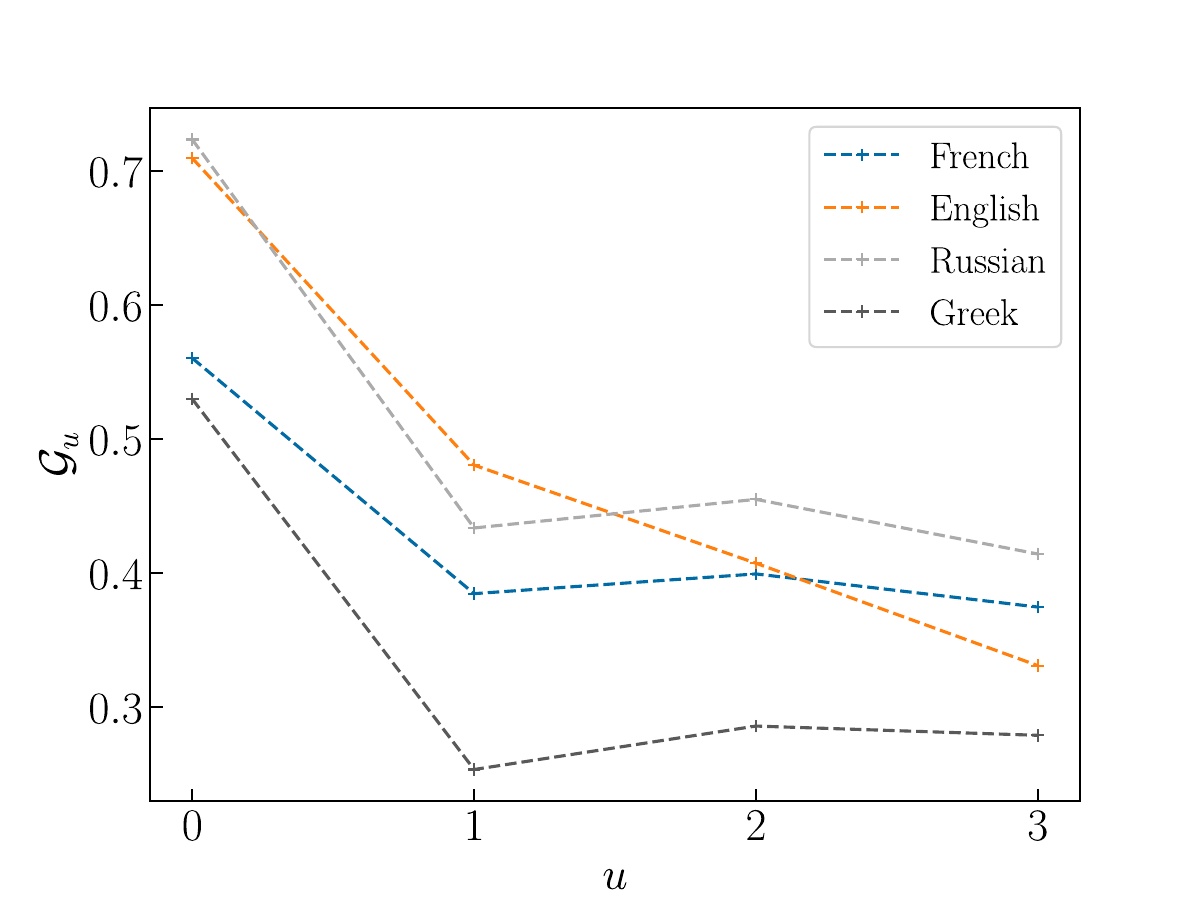}
\caption{Predictability gain $\mathcal{G}_u$ as a function of the phonological block size for French (for which the phoneme inventory size is $L = 35$), English ($L = 38$), Russian ($L = 47$) and Greek ($L = 30$).}
\label{predgains}
\end{figure}

\begin{figure}[t]
    \centering
    \includegraphics[width=.45\textwidth,clip]{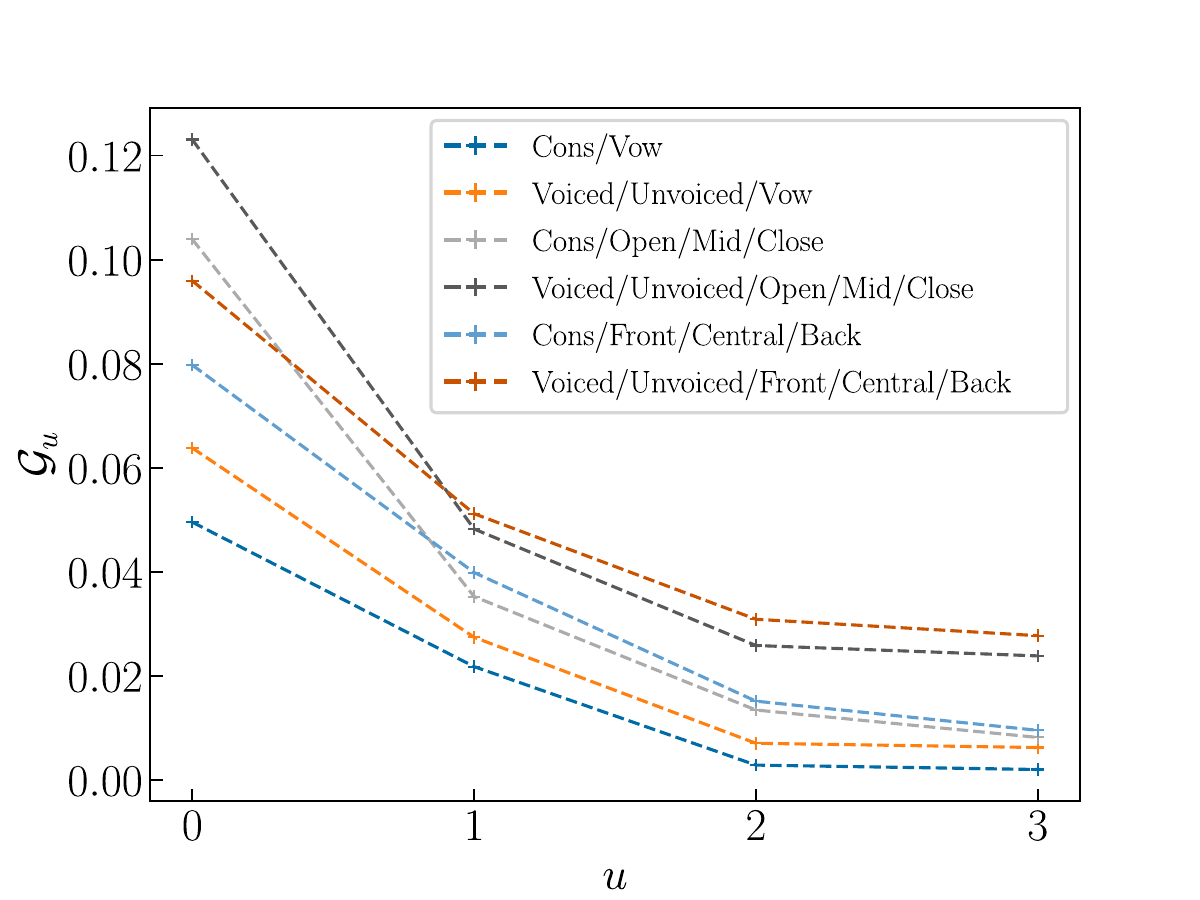}
        \includegraphics[width=.45\textwidth,clip]{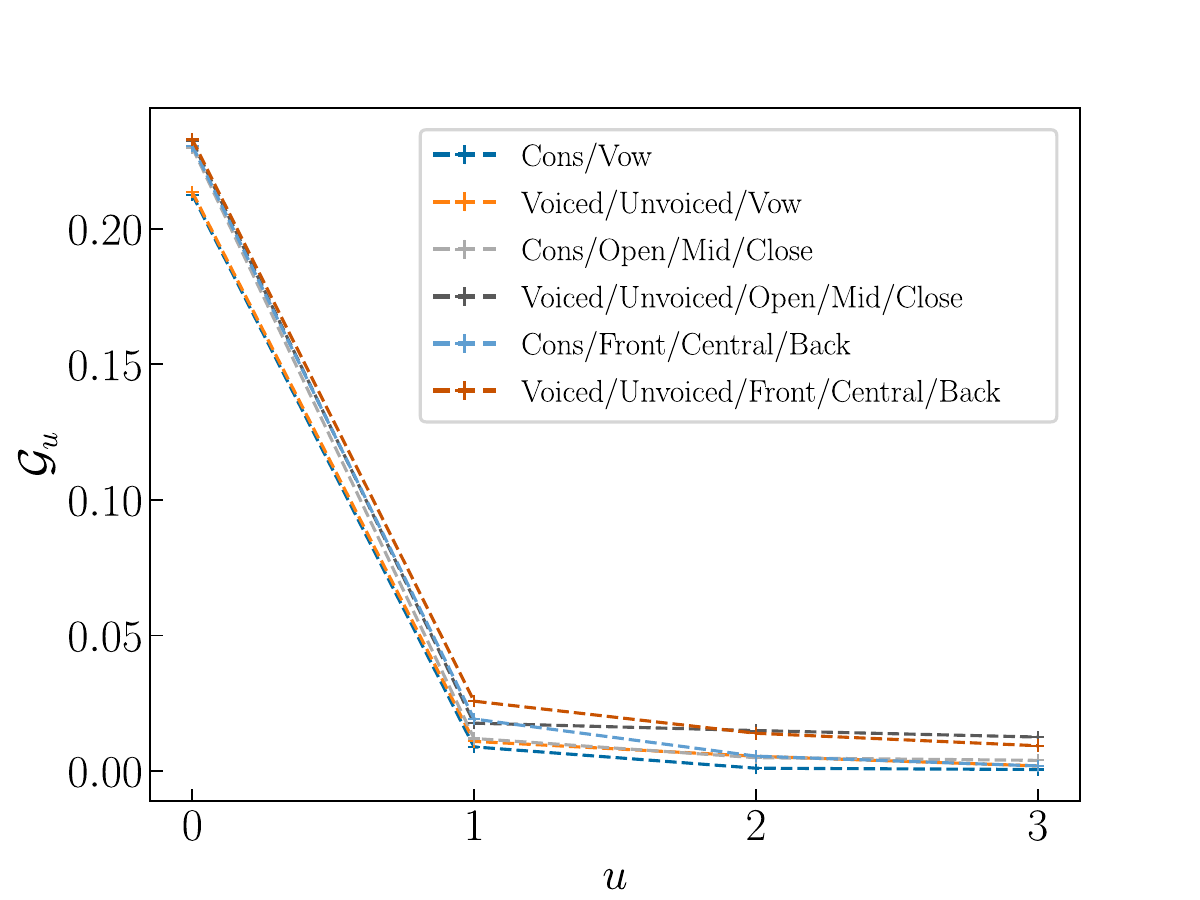}
\caption{Left: Predictability gain for English phonological classes as a function
of the block size. Consonants can be either all grouped together (Cons) or divided into voiced and unvoiced.
Vowels can be grouped in one class (Vow) or classified in terms of openness (open, mid, or close) or backness (front, central or back).
The value of $L$ for each curve is the number of categories considered therein (i.e., $L=2$ for the Cons/Vow curve, $L=3$ for the Voiced/Unvoiced/Vowel curve, etc.).
Right: Same as the left panel but for the French language.}
\label{predgains2}
\end{figure}

Thus, we compute the first 4~values of $\mathcal{G}_u$ for all languages in our dataset, four of which are displayed in Fig.~\ref{predgains}. We observe that the predictability gain for $u=0$ is larger than for $u=1$, implying that information is gained when shifting from a description of purely independent phonemes to a classical Markovian model based on first-order transition probabilities. As $u$ increases, we arrive at an approximately steady behavior
of $\mathcal{G}_u$ for $1\le u\le 3$. To improve the reliability of $\mathcal{G}_u$ around $u=3$ we need
to increase $r_{\rm max}$. This can be done by
a coarse-graining approach that consists of grouping consonants by voicing (voiced or voiceless) and vowels by either openness (open, mid or close) or backness (front, central or back). Therefore, $L$ reduces to values
between 2 and 5 and $r_{\rm max}$ consequently increases.
Then, we compute the corresponding probability distributions and block entropies associated to these
new $r$-phone classes. We present the predictability gain in Fig.~\ref{predgains2} for English (left panel) and French (right panel). We find a strong decrease between $u=0$ and $u=1$ and then a more uniform behavior for higher values of $u$, in agreement with Fig.~\ref{predgains}.
In fact, for $L=2$ we obtain $\mathcal{G}_u\simeq 0$ at $u=3$.
Taken together the results
of Figs.~\ref{predgains} and~\ref{predgains2}, we can safely estimate the memory to be~$2$ since values $u\ge 3$
do not yield any additional significant information. Of course, the true generative process may involve longer-range correlations, but we consider that a memory $m=2$ is the best compromise between statistical significance and computation reliability.
As a consequence, 3-phone models capture, to a very good extent, the essential statistical properties of phonological variation in our corpus. We note in passing that triphone speaker models
have been successfully applied in speech recognition~\cite{kohler2001phonetic}
and trigrams have been found to accurately describe parts-of-speech sequences~\cite{de_gregorio_exploring_2024}.

\subsection{Distance metric}
\begin{figure}[t]
    \centering
    \includegraphics[width=.45\textwidth,clip]{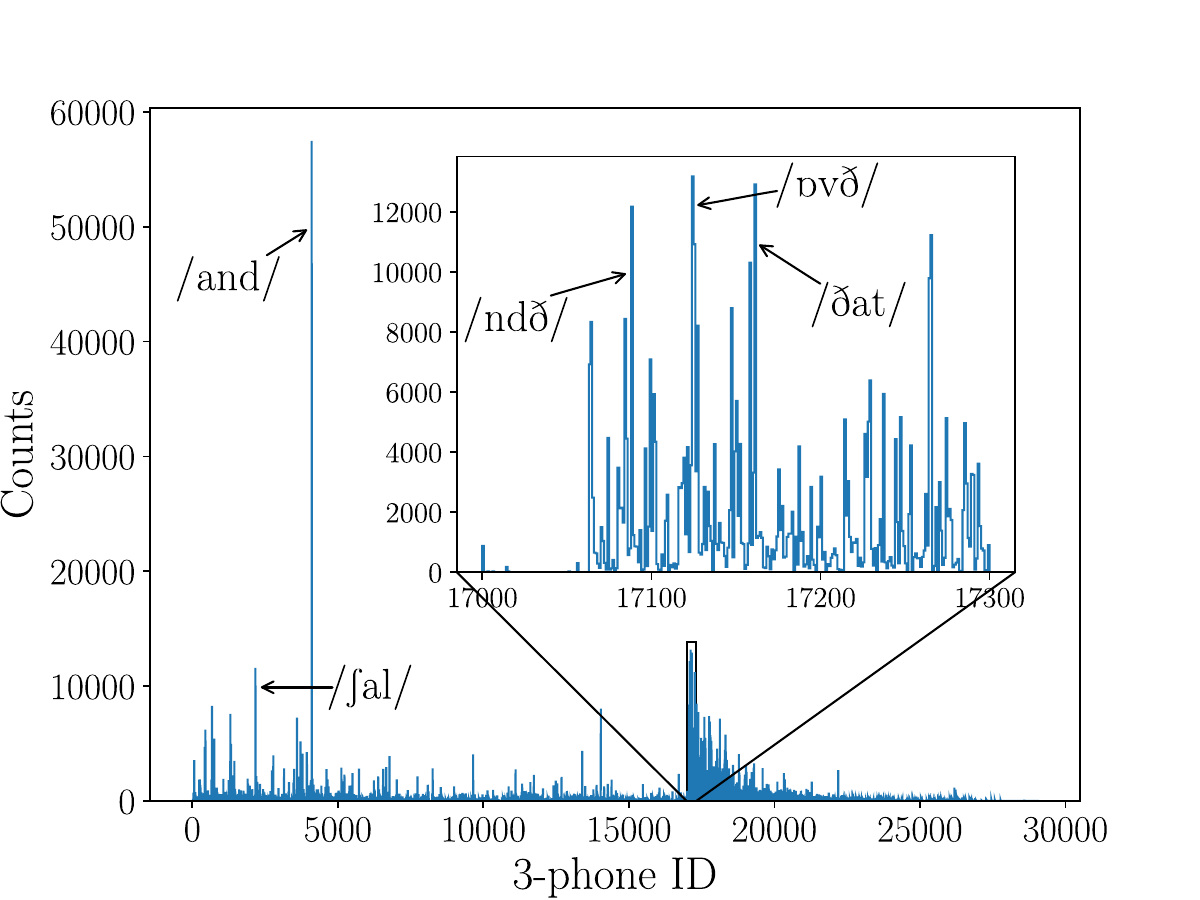}
        \includegraphics[width=.45\textwidth,clip]{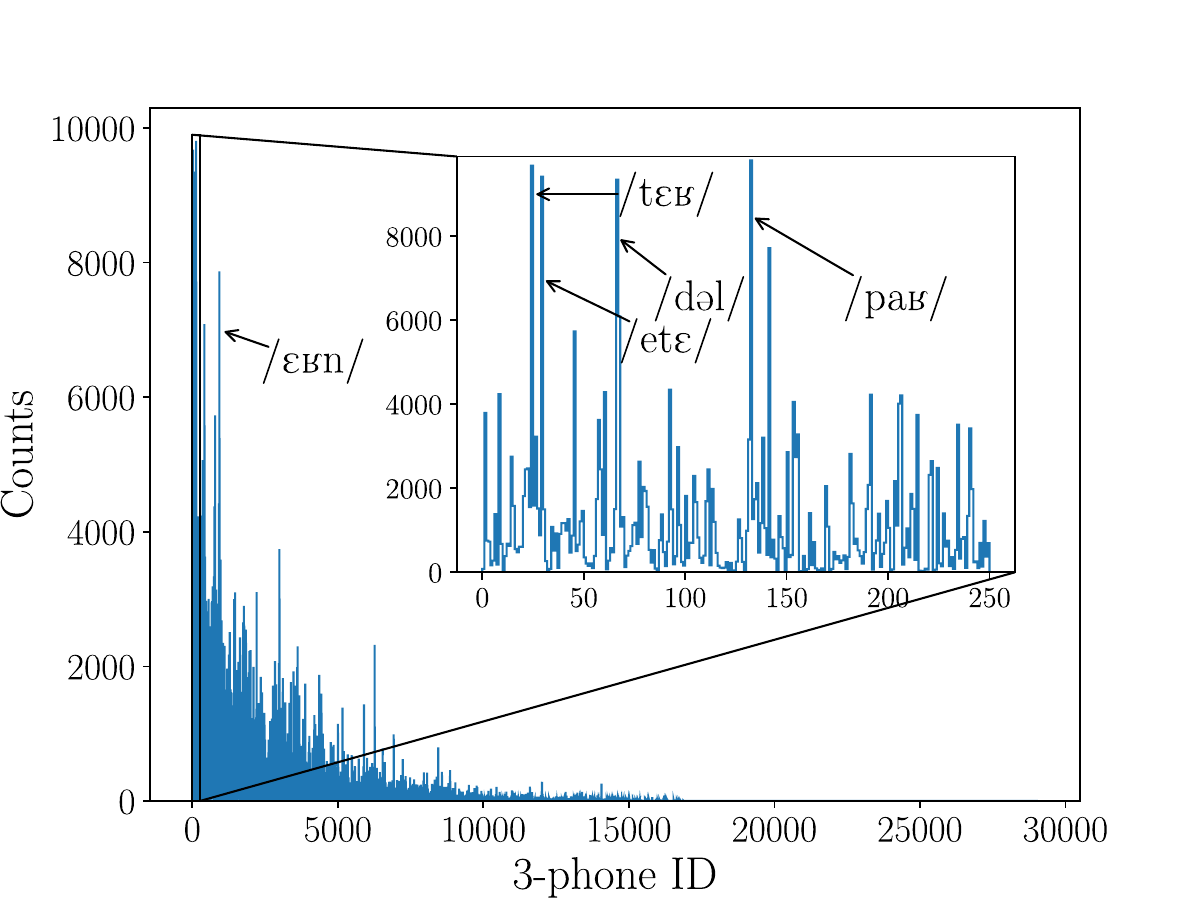}
\caption{Left: Distributions of English phoneme trigrams. The five most frequent 3-phones are indicated with arrows.
Right: Same for French.}
\label{distrib}
\end{figure}

Our goal is now to define the phonological distance between two languages that quantifies the divergence between their respective phonological systems characterized by the 3-phone model. As an illustrative example, we show in Fig.~\ref{distrib} the 3-phone distributions for English (left panel) and French (right panel). A few phoneme trigrams are clearly interpretable, such as /\textipa{and}/, mostly corresponding to the conjunction \emph{and}, and /\textipa{Sal}/, which reflects the auxiliary \emph{shall}, both of which are high-frequency terms in our biblical corpora.

However, if one were to use a divergence measure based solely on the
occurrence frequencies such as those in Fig.~\ref{distrib}, all 3-phones would effectively be treated as equally distant. This assumption is linguistically
unrealistic, since many triphones share substantial phonological
similarity. For instance, the 3-phones /\textipa{buk}/ and /\textipa{bux}/
(corresponding to the cognate words \emph{book} and \emph{Buch} in English
and German) are clearly more similar to each other than to the vast majority of
3-phones. To capture this relevant structure,
we put forward a feature-based embedding
that reflects phonological similarity between segments. We represent each phoneme as a vector of 24~articulatory features~\cite{mortensen_panphon_2016}: syllabic, sonorant, consonantal, continuant, delayed release, lateral, nasal, strident, voice, spread glottis, constricted glottis, anterior, coronal, distributed, labial, high (vowel/consonant, not tone), low (vowel/consonant, not tone), back, round, tense, and 4 features which do not apply here (velaric (click), long, and 2 tonal features). Each feature takes one of the following values: $+1$ if the feature is present, $-1$ if the feature is absent or $0$ if the feature is unspecified, i.e., irrelevant for that phoneme. For instance,
the feature `anterior', which describes segments
produced with a constriction in the front of the vocal tract, yields $+1$ for /\textipa{t}/, $-1$ for /\textipa{S}/ and 0 for /\textipa{a}/ because this feature is applicable only to consonants.  As a consequence, after concatenation of three feature vectors, every 3-phone is mapped to a 72-dimensional vector, which in practice reduces to 60 dimensions, due to the 4 features that are constant over the dataset.

Let $d(x_i,y_j)$ be the distance between the 3-phone vectors $x_i$ and
$y_j$. We calculate $d$ from the feature edit distance, which assigns substitution costs according to similarity between $x_i$ and
$y_j$ in feature space~\cite{mortensen_panphon_2016}. In Appendix~\ref{app_alt}, we perform an analysis of alternative distances that further supports our choice for this metric. Once the ground metric is fixed, we define the phonological distance between language $\mathcal{L}$
and language $\mathcal{L}'$ by means of the Wasserstein (or Earth mover's) distance~\cite{kantorovich_mathematical_1960,panaretos2019statistical},
which has been applied in fields as diverse as
computer vision~\cite{rubner1998metric,levina},
genomics~\cite{bolstad2003comparison,evans2012phylogenetic}
or even linguistics~\cite{alvarez2018gromov,louf2021capturing}:
\begin{align}
    W(\mathcal{L}, \mathcal{L}') = \min_{\gamma \in \Gamma(P, Q)} \sum_{i} \sum_{j} \gamma_{ij} d(x_i, y_j) \,.\label{eq_W}
\end{align}
Here, $P$ ($Q$) is the probability distribution for 3-phones $x_i$ ($y_j$) in language $\mathcal{L}$  ($\mathcal{L}'$) that occur with frequency $P(x_i)$
$[Q(y_j)]$. The minimization in Eq.~\eqref{eq_W} is carried over the set
$\Gamma(P,Q) =\{ \gamma_{ij} \ge 0 :
\sum_{j} \gamma_{ij} = P(x_i) \ \forall i,\,\,
\sum_{i} \gamma_{ij} = Q(y_j) \ \forall j \}$,
where, as indicated, the variables $\gamma_{ij}$ are subject to the constraints that their marginals must equal $P$ and $Q$. Therefore,
this procedure measures the minimum cost of transforming
$P$ into $Q$ under the articulatory-feature–-based metric $d$
and is consequently a proper measure of the phonological divergence
between two languages.

\section{Results and Discussion}

\subsection{Phonological distances}

\begin{figure}
    \centering
    \includegraphics[width=.95\textwidth,clip]{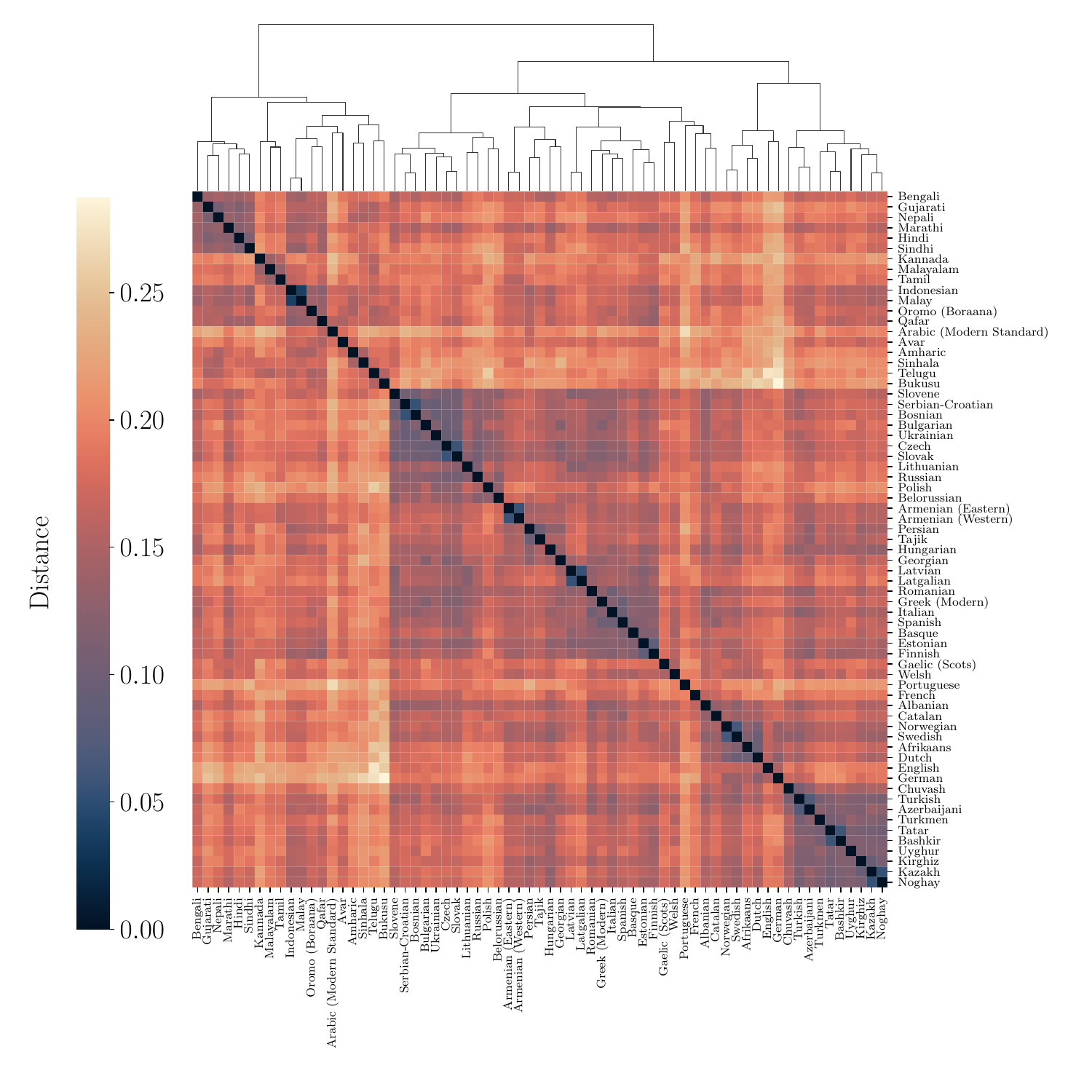}
\caption{Heatmap representation of the phonological distances between all pairs of languages in our sample, calculated on the 3-phone probability distributions under a metric based on articulatory feature vectors. Languages are ordered according to the clusters found by hierarchical clustering using the Ward linkage, visualized as a hierarchical tree.}
\label{all_WS_clustermap}
\end{figure}

We compute $W(\mathcal{L}, \mathcal{L}')$ for all language pairs, using an approximation based on the Sinkhorn algorithm~\cite{cuturi2013sinkhorn}, which introduces an entropic regularization term to the linear optimal transport problem~\cite{flamary_pot_2021}. Thus, we construct
the phonemic distance matrix $M$ among the 67 languages in our dataset. We display $M$ in Fig.~\ref{all_WS_clustermap}
along with the hierarchical clusters found
by applying Ward linkage~\cite{ward1963hierarchical}.

While $M$ reveals clear clustering patterns among languages, we emphasize that these groupings reflect sound-based relatedness, not necessarily genetic affiliation. Although languages from the same family often exhibit similar phonological profiles, this is not a strict rule. Languages may cluster phonologically due to contact, convergence or shared phonemic tendencies independent of genealogical descent. Therefore, our following discussion focuses on phonological proximity, without assuming a necessary correspondence with genetic classification.

First, all Altaic languages of our sample have been grouped together in a well-defined cluster at the bottom-right corner of Fig.~\ref{all_WS_clustermap}.
Despite the existence of the Altaic family is debated, we prefer to keep it for consistency with the World Atlas of Language Structures Online (WALS)~\cite{dryer_wals_2013}, which we use below.
Another clear groups comprise, on the one hand, the Germanic languages and, on the other hand, the Slavic genus, to which Lithuanian is added. This is not surprising since Baltic and Slavic languages are often considered to be part of a larger branch of the Indo-European family termed Balto-Slavic~\cite{pronk202215}. Yet, one should note that the other Baltic languages, Latvian and Latgalian, are not classified in that cluster, but seem to appear closer to a few Romance languages, which are correctly grouped together. This might be an artifact due to the closeness between Latvian and Latgalian, with high mutual intelligibility.

Next, the Romance family is disseminated in two clusters, the first one being composed of Romanian, Italian and Spanish, to which Greek is added, with a subcluster featuring Basque, Estonian and Finnish. The similarity between Spanish and Basque phonologies could be due to the long-lasting contact between these two languages~\cite{hualde}. Greek and Spanish also share plenty of common phonemes, which include most of their vowels, as well as consonants /\textipa{G}/, /\textipa{D}/, /\textipa{T}/ and /\textipa{x}/ which are rather rare throughout the sample languages. Even though the resemblance between Finnic and Romance languages is less documented, the vowel inventories of Spanish and Estonian, for instance, are rather similar~\cite{leppik2014comparative}, which could explain part of the clustering. The other Romance languages (French, Portuguese and Catalan) are grouped in another cluster, along Albanian. This is expected since French and Portuguese differ from other Romance languages by, e.g,, the presence of nasal vowels. Moreover, Catalan is considered to be closer to French than to Spanish, in terms of lexicon, grammar as well as phonology~\cite{feldhausen2010sentential}.
On the other hand, the phonological proximity of Albanian with the Romance languages may stem from a long contact period with Late Latin~\cite{mallory_albanian_1997}.

Then, we find a cluster in the top-left corner of Fig.~\ref{all_WS_clustermap}, regrouping almost all of the Indo-Aryan languages of our sample, which indeed share a good deal of phonological features, including aspirated or retroflex consonants which are almost exclusive to these languages throughout the sample. The only exception is Sinhala, an Indo-Aryan language spoken in Sri Lanka that lacks aspiration in modern times, and is consequently regrouped with Amharic in another cluster. 
 
Further, we obtain interesting small groups including the Western and Eastern forms of Armenian, which are located near the Persian languages (Farsi and Tajik) in the cluster tree, a proximity that may be influenced by their geographic closeness. Three of the four Dravidian languages of our sample (Kannada, Malayalam and Tamil) are in a cluster adjacent to the Indo-Aryan genus, as well as Indonesian and Malay, which are the only members of the Austronesian language family in our data. This relatedness between languages from different families can also be explained by contact. In particular, contacts between Indo-Aryan and Dravidian languages have been the object of numerous studies and are, e.g., considered to be responsible for the introduction of retroflexes in Indo-Aryan~\cite{tikkanen2003archaeological}. 

\begin{figure}[t]
    \centering
    \includegraphics[width=.95\textwidth,clip]{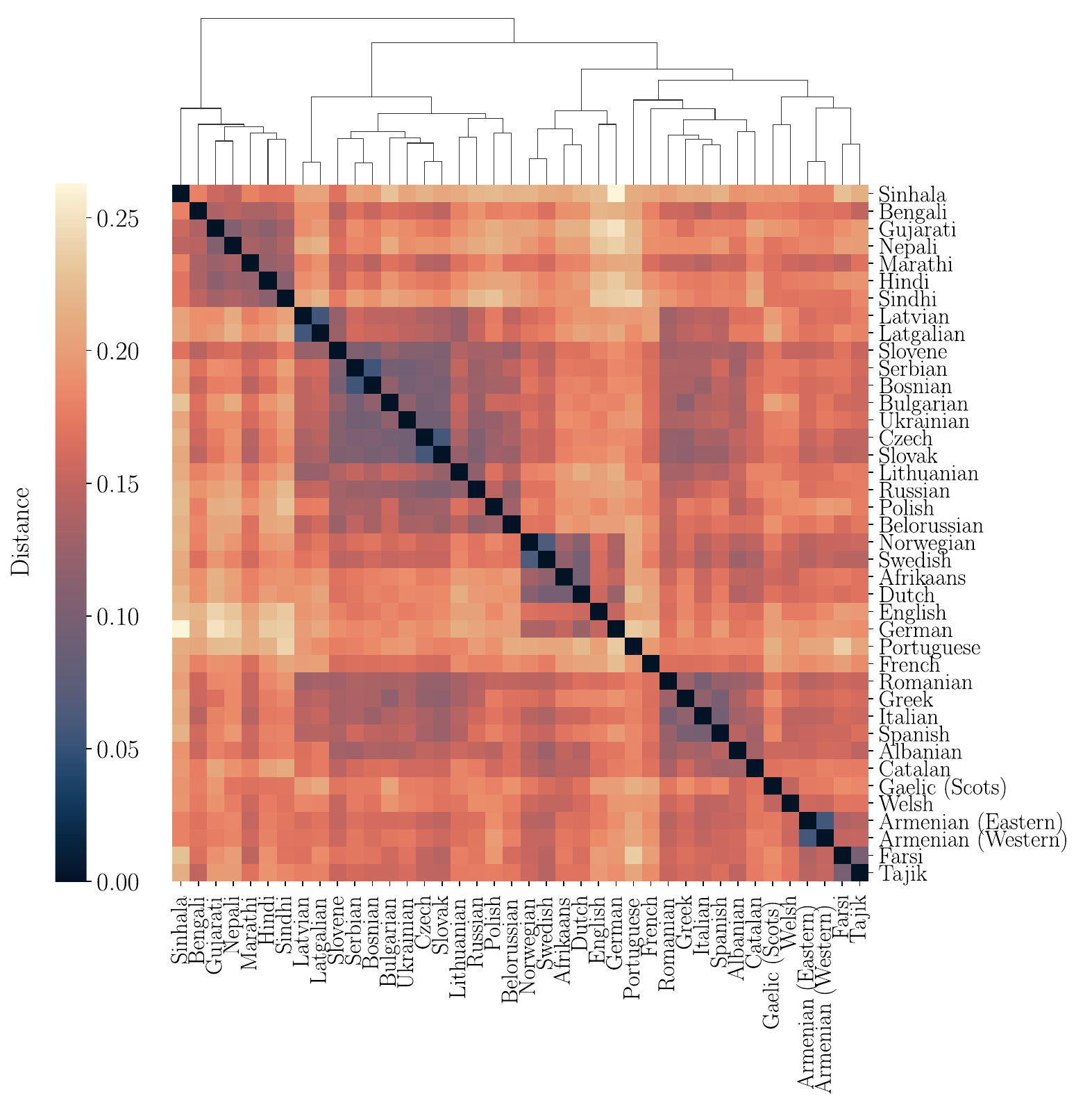}
\caption{Same heatmap as Fig.~\ref{all_WS_clustermap} but showing only Indo-European languages Languages. The hierarchical clustering method is now applied to this family, using the Ward linkage, and is displayed as a hierarchical tree. We also order languages according to the clusters.}
\label{IE_WS_clustermap}
\end{figure}

Let us now focus on the 40~Indo-European (IE) languages, which form the largest language family present in our dataset.
We present the results in Fig.~\ref{IE_WS_clustermap}. This time, Sinhala is correctly grouped with the other Indo-Aryan languages. Across the diagonal,
we find the rest of the major IE sub-families, namely,
Balto-Slavic, Germanic and Italic. Since the Albanoid and Hellenic sub-families have one language each in our dataset, they are clustered in the larger group of Romance languages for the same reasons that we invoked while interpreting Fig.~\ref{all_WS_clustermap}. Overall,
we do not find significant differences between
the clusters in Figs.~\ref{all_WS_clustermap} and~\ref{IE_WS_clustermap}, suggesting that phonological similarities between IE languages are strong enough to dominate clustering even when we add other language families to the analysis~\footnote{Incidentally, we can employ a phonological distance calculation to quantitatively verify that our corpus is representative. Thus, we compute the distance between English probability distributions of the Bible and Herman Melville's Moby Dick, a frequently analyzed text in computational and quantitative linguistics
[W.~Ebeling and T.~Poschel, Europhysics Letters {\bf 26}, 241 (1994); C.~Akimushkin, D.~R.~Amancio, and O.~N.~Oliveira Jr, PLOS ONE {\bf 12}, e0170527 (2017)]. We find a distance of $0.0538$ between both texts, which is lower than the smallest distances in our sample (between Latvian and Latgalian, as discussed earlier). Considering that the two books have distinct vocabulary, writing styles, were published 200~years apart and that Moby Dick is only half as long as the Bible, this test demonstrates that the representativeness of our corpus suffices for the purposes of our work.}.

\subsection{Connection with geographical distances}
As pointed out in Sec.~\ref{sec:intro}, recent data-driven approaches~\cite{jeszenszky2017exploring,jager2018global,wichmann2010homelands,de2024exploring} have shown a robust positive correlation between geographic proximity and linguistic similarity. This is expected on average and is mainly due to two reasons: shared phylogenetic inheritance, whereby languages spoken in nearby regions often descend from a common ancestor, and contact-induced change, through which neighboring languages influence one another over time. As geographic distance increases, both genealogical ties and interaction likelihood decrease and we expect a greater divergence across different linguistic domains. Our aim is now to investigate whether this correlation is observed in the phonological domain.

\begin{figure}[t]
     \centering
        \includegraphics[width=.45\textwidth,clip]{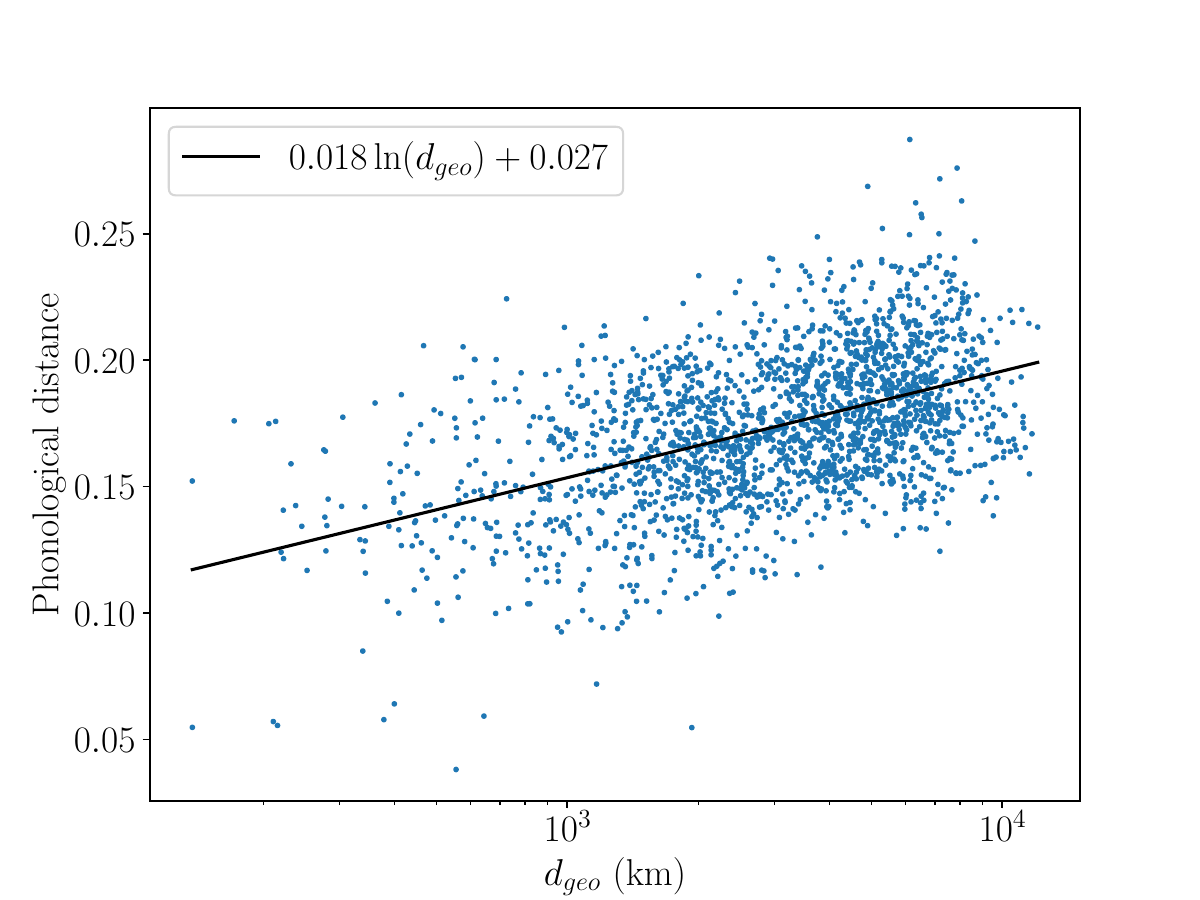}
        \includegraphics[width=.45\textwidth,clip]{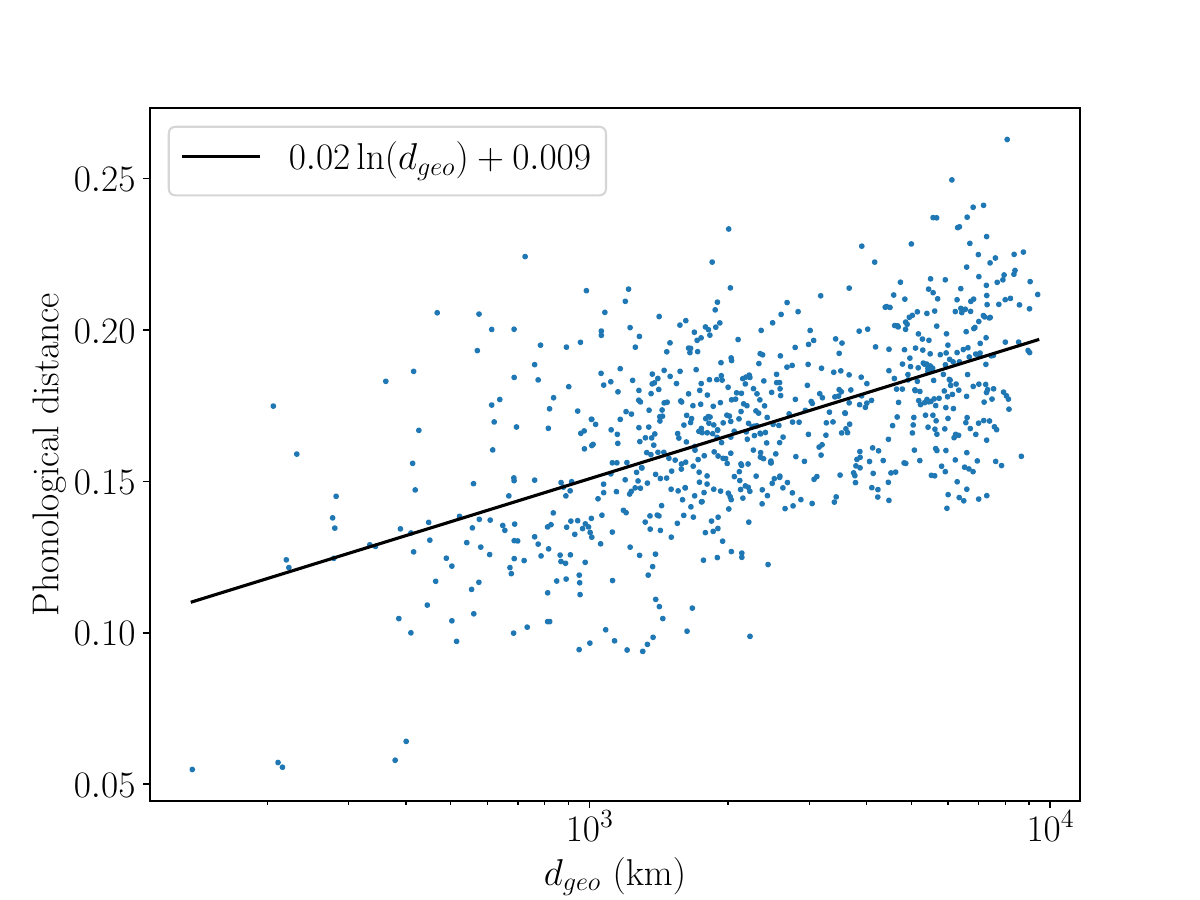}    
        \caption{Wasserstein distance between the 3-phone probability distributions of (left) all languages pairs and (right) IE language pairs only, versus geographic distance $d_{geo}$. On each plot, a logarithmic fit is shown in black.}
        \label{phongeocor}
\end{figure}

To this end, we assign WALS geographic coordinates to each language~\cite{dryer_wals_2013} and compute the geodesic distance $d_{geo}$ on Earth's surface between these coordinates. We exclude Afrikaans from this analysis since this language is the only product of transcontinental colonization in our dataset. Then, we plot the phonological distances calculated from the Wasserstein metric as a function of $d_{geo}$ shown in logarithmic scale. We present the results in Fig.~\ref{phongeocor} for all languages in our dataset (left panel) and for IE languages only (right panel). In both cases, we indeed observe a correlation pattern that
points to a greater phonological distance as $d_{geo}$ grows.
We find that this pattern is well fitted by a logarithmic function, as shown in Fig.~\ref{phongeocor}.

The correlation is stronger when we consider IE languages only,
as evidenced by a calculation of the distance correlation coefficient $R_d$~\cite{szekely2007measuring}, which generalizes Pearson's coefficient to non-linear relations. We obtain $R_d = 0.496$ for IE languages and
$R_d = 0.428$ for all languages, with p-values $<0.001$ calculated through permutation tests. This suggests that phylogeny plays a non-negligible role in the phonological distances, which needs to be disentangled from mere geographic contact. We next use this finding as the building block for a method aimed at identifying the homeland of the IE family.

We now calculate the average 3-phone probability distribution $P_{\rm av}$ of the 39~IE languages of our dataset (we recall that Afrikaans is excluded from the original list). Note that, instead of aggregating all phoneme occurrences in our corpus and then computing their average, we average the individual distributions since each language has its own phonological system. We emphasize that $P_{\rm av}$ does not correspond by any means to the 3-phone distribution of the Proto-Indo-European language. Rather, $P_{\rm av}$ represents the average phonemic diversity of the IE family.
Since languages within the same family will tend to diverge phonetically as their speakers migrate away from the homeland, it is reasonable to expect that the phonological distance $d_{p_i}$ between language $i$ and $P_{\rm av}$ will increase with the geographical distance. A similar argument has been employed to investigate the phonetic diversity in world languages
from a series of founder effects~\cite{atkinson2011phonemic,fort2016can}. Naturally, as discussed above, this divergence is also influenced by language contact~\cite{jaeger2011mixed, hunley2012rejection}. However, we argue that the phylogenetic signal in our data is sufficiently strong, primarily because our analysis is confined to a single language family rather than spanning an entire set of world languages, and also because
we consider the full phoneme frequency distributions, not only their presence or absence.

\begin{figure}[t]
    \centering
    \includegraphics[width=1.05\textwidth,clip]{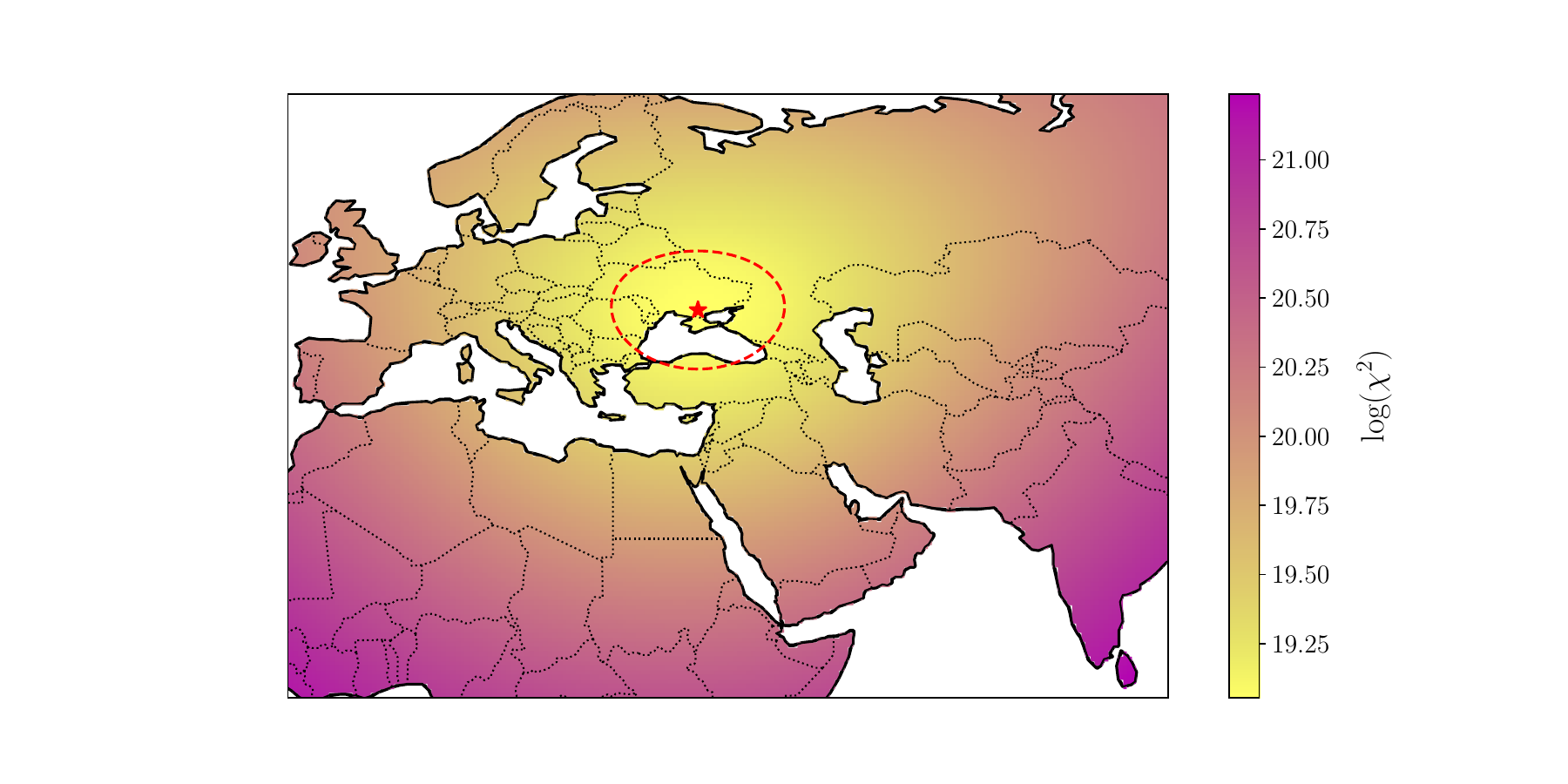}
    \caption{Heatmap for the sum of squared residuals $\chi^2$ between the actual and fitted geographical distances that correspond, respectively, to Indo-European language locations and potential family origins. The point with the smallest residual, and consequently the most likely homeland for the Indo-European family, is indicated with a star. We depict its 95\% uncertainty region with a dashed circle, calculated from $K=2000$ random samples of the weights associated with the languages.}
    \label{mapkhi2}
\end{figure}

We compute $d_{p_i}$ using Eq.~\eqref{eq_W} and employ the fit in the right panel of Fig.~\ref{phongeocor} to transform $d_{p_i}$ into a geographical distance $d_{g_i}$. Thus, for any point on Earth with geographical coordinates $r$ at distance ${d}(r_i,r)$ of language $i$ with WALS coordinates $r_i$, we define the sum of squared residuals
\begin{align} \label{eq_chi2}
    \chi^2 (r) = \sum_{i=1}^{39} \left[d_{g_i} - {d}(r_i,r)\right]^2\,,
\end{align}
This function measures the total discrepancy between the geographic distances predicted from phonological divergence and the actual geographic distances to a candidate location. Minimizing $ \chi$ therefore identifies the geographic points from which the pattern of phonological divergence across IE languages is most consistent with geographic distance. Under the previously stated assumption that phonological distance tends to increase with separation from the ancestral population, the location with minimum $\chi$,
 \begin{align}
     r^* = \arg\min_r \chi^2 (r)
 \end{align}
can be interpreted as a plausible candidate for the IE homeland. In Fig.~\ref{mapkhi2} we show the results for $\chi^2$ along with $r^*$.
We note that minimizing Eq.~\eqref{eq_chi2} amounts to minimizing the average
$\chi^2_{av}=\chi^2/39$, which corresponds to assigning equal weights to all languages. To define an uncertainty region around $r^*$, we assign weights $w_1,\ldots,w_{39}$ randomly drawn from a Dirichlet distribution with all concentration parameters equal to 1. This ensures that the weights are non-negative, add up to one and each have expected value equal to their original weight. Moreover, the joint probability density function for $w_1,\ldots,w_{39}$ is uniform~\cite{balakrishnan2004primer}. Hence, we define
 \begin{align} \label{eq_chi22}
    \chi_{av}^2 = \sum_{i=1}^{39} {w}_i \left[d_{g_i} - {d}(r_i,r)\right]^2\,.
\end{align}
Repeating this procedure $K$ times, we obtain the locations $r_1^*,\ldots,r_K^*$,
corresponding to the values that minimize Eq.~\eqref{eq_chi22} for each draw of the weights.
We calculate ${d}(r^*,r^*_k)$ for $k=1,\ldots, K$ and define $R$ as the 95th
percentile of this distribution. We thus obtain the dashed circle centered
at $r^*$ and radius $R$ as illustrated in Fig.~\ref{mapkhi2}.

 Notably, $r^*$ and its 95\% uncertainty region lie mostly north of the Black Sea, a result which is consistent with the Kurgan hypothesis that IE languages originated in the Pontic steppe~\cite{gimbutas2022old,mallory,chang2015ancestry,kroonen2022indo}.
 This is also in agreement with very recent findings from genetics~\cite{lazaridis2025genetic}.
 Our results may be in less good agreement, but not entirely incompatible, with the Anatolian hypothesis, which advocates an earlier origin in Anatolia~\cite{renfrew1,bouckaert2012mapping}.

\section{Conclusions}
We have shown that phoneme sequences of 67 different modern languages can be efficiently modeled as 2-order Markov chains. This enables us to investigate interlinguistic phonological similarities using a metric suitable for triphone probability distributions. Accordingly, we have recovered well known language groups as well as non genetically-driven similarities that can be explained by language contact. A limitation of our approach is the size of the dataset under consideration. More languages could be included to the analysis to capture more accurately the diversity and complexity of their phonological relationships, especially for non-IE languages. One could also consider varieties of a given language and investigate further interesting problems in sociolinguistics related to socially conditioned pronunciation patterns~\cite{labov}.

We have then found a significant correlation between phonological distance and geographic proximity. This result relies on assigning unique geographic coordinates to each language, a limitation that could be overcome by instead measuring distances between linguistic areas rather than points~\cite{haspelmath2001european,masica2005defining}.
We have exploited the obtained correlation to constrain a potential origin region for the Indo-European family.
Our main result is consistent with the Steppe hypothesis. Here, a promising avenue would be to adapt the approach used in Ref.~\cite{cysouw2013chapter} to explore the link between genealogical and geographical distance.
Although our method neglects spatial autocorrelation,
this could be fixed by modeling residuals as spatially-dependent. However, the minimization procedure would become cumbersome and can be postponed for a future work.

Finally, we remark that our approach is synchronic. One could add the time axis to our phonological matrix distance if a comparable multilingual, diachronic corpus were available. A long time depth would be essential
if one wishes to address language evolution phenomena. Alternatively, one could apply Bayesian phylogenetic methods~\cite{greenhill2020bayesian}
to estimate dates of phonological divergences.

\begin{acknowledgments}
This work was supported by Grant
No.\ PID2024-157493NB-C21/C22 funded by MICIU/AEI/10.13039/501100011033 and by ``ERDF/EU'', and the Mar\'{\i}a de Maeztu Program for Units of Excellence
in R\&D, Grant No.\ CEX2021-001164-M.
\end{acknowledgments}

\appendix

\section{List of languages} \label{app_data}

The next three tables present the list of languages in our corpus, with basic information on phylogeny, geography and the source text. Unless otherwise specified, geographical coordinates were extracted from the WALS database and speaker numbers from Ethnologue~\cite{ethnologue15}. Languages transcribed with Epitran are marked with \textsuperscript{\textdagger}, while the rest are transcribed using Phonemizer. Languages whose Bible translations are gathered on Ref.~\cite{biblecom} are written in bold, whereas the rest are collected from Ref.~\cite{christodouloupoulos2015massively}.

\newpage
\clearpage
\begin{sidewaystable} 
\begin{longtable}{ |c|c|c|c|c|c|c|c|c|c| } 
\hline
\rowcolor{lightgray}  Language family & Genus & Language & Speakers &Latitude & Longitude & Script & Year & Words & Phonemes\\
\hline
\multirow{4}{5em}{Afro-Asiatic} & \multirow{2}{6em}{Lowland East Cushitic} &\textbf{Oromo (Boraana)\textsuperscript{\textdagger}} & 3,827,616 & 4.50 & 38.50 & Latin & - & 527,105 & 2,740,241 \\
&& \textbf{Qafar}\textsuperscript{\textdagger} & 1,439,367 & 12.00 & 42.00 & Latin & - & 311,094 & 1,570,447\\\cline{2-10}
&\multirow{2}{4em}{Semitic} & Amharic & 17,417,913 & 10.00 & 38.00 & Ethiopic & 1840 & 414,805 & 3,135,075 \\
& & Arabic & 206,000,000 & 25.00 & 42.00 & Arabic & 1865 & 435,213 & 2,279,355 \\
\hline
\multirow{10}{3em}{Altaic} & \multirow{10}{3em}{Turkic} & \textbf{Azerbaijani}& 7,059,529 & 40.50 & 48.50 & Latin & 2008 & 475,781 & 2,801,455 \\
&& \textbf{Bashkir} & 1,871,383 & 53.00 & 58.00 & Cyrillic & 2023 & 366,023 & 2,133,947 \\
&& \textbf{Chuvash}& 1,834,384 &55.50	&47.50& Cyrillic & - & 133,499 & 752,794 \\
&& \textbf{Kazakh} & 8,178,879 & 50.00 & 70.00 & Cyrillic & 2010 & 517,197 & 3,170,171\\
&& \textbf{Kirghiz} & 3,136,733 & 42.00 & 75.00 & Cyrillic & 2005 & 188,508 & 1,173,104 \\
&& \textbf{Noghay} & 67,806 & 44.00 & 46.00 & Cyrillic & 2011 & 113,701 & 661,645 \\
&& \textbf{Tatar} & 1,610,032 & 56.00 & 49.50 & Cyrillic & 2015 &514,785& 2,993,176  \\
&& \textbf{Turkmen} & 6,403,533 & 40.00	& 58.00 & Latin & 2016 & 389,079 & 2,307,694 \\
&& Turkish & 50,625,794 & 39.00 & 35.00 & Latin & 1827 & 454,119 & 2,860,584 \\
&& \textbf{Uyghur} &7,601,431& 40.00 & 80.00 & Arabic & 2005 & 539,711 & 3,513,554 \\
\hline
\multirow{2}{6em}{Austronesian} & \multirow{2}{6em}{Malayo-Polynesian} & Indonesian & 23,143,354 &0.00	&106.00& Latin & 1974 & 659,540 & 3,578,631 \\
&& \textbf{Malay} & 17,604,253 & 3.00 & 102.00 & Latin & 1996 & 621,694 & 3,434,690 \\

\hline
Basque & Basque &  Basque & 588,108 &43.00	&-3.00& Latin & 1855 & 133,761 & 813,563 \\
\hline
\multirow{4}{4em}{Dravidian} & \multirow{4}{4em}{Dravidian} & Kannada & 35,346,000 &14.00 & 76.00& Kannada & 1831 & 492,653 & 3,817,761 \\ 
&& Malayalam & 35,757,100 & 10.00 & 76.50 & Malayalam & 1841 &430,356 & 1,381,126 \\ 
&& \textbf{Tamil} & 66,020,200& 11.00 & 78.50 & Tamil & - & 352,148 & 1,340,475\\
&& Telugu & 69,688,278 & 16.00 & 79.00& Telugu & 1854 & 443,045 & 3,084,751 \\ 
\hline
Kartvelian & Kartvelian & \textbf{Georgian} & 4,178,604 & 42.00	& 44.00 & Mkhedruli & 2002 & 467,018 & 738,154 \\ \hline
\hline

\end{longtable}
\end{sidewaystable}

\newpage
\begin{sidewaystable}    
\begin{longtable}{ |c|c|c|c|c|c|c|c|c|c| }
\hline
\rowcolor{lightgray} Language family & Genus & Language & Speakers &Latitude & Longitude & Script & Year & Words & Phonemes\\
\hline
\multirow{22}{6em}{Indo-European} & Albanian &  Albanian & 2,980,000 & 41.00 & 20.00& Latin & 1993 & 750,035 & 3,004,925 \\ \cline{2-10}
&\multirow{2}{4em}{Armenian} & Armenian (Eastern) & 6,723,840 &40.00&45.00& Armenian & - & 534,548 & 3,013,350  \\
&& \textbf{Armenian (Western)} &1,600,000 & 38.50&43.50& Armenian & 1853 & 475,887 & 2,447,648 \\ \cline{2-10}
 & \multirow{3}{3em}{Baltic} & \textbf{Latgalian} & 200,000\ & 56.00 & 27.00 & Latin & 1937& 135,198 & 692,819 \\
&& Latvian & 1,543,844 & 57.00 & 24.00& Latin & 1689 & 132,517 & 675,583 \\
&& Lithuanian & 3,125,281 & 55.00 & 24.00& Latin & 1735 & 469,601 & 2,347,227 \\ \cline{2-10}
& \multirow{2}{3em}{Celtic} & \textbf{Gaelic (Scots)}& 62,175 & 57.00 & -4.00 & Latin & - & 549,816 & 1,896,731\\
&& \textbf{Welsh} & 536,258 & 52.00 & -3.00 & Latin & 1588 & 668,413 & 2,504,930 \\ \cline{2-10}
&\multirow{6}{4em}{Germanic} &  Afrikaans & 5,865,879 &-31.00 & 22.00& Latin & 1953 & 799,655 & 3,019,921 \\
&& Dutch & 17,370,777 & 52.50 & 6.00 & Latin & - & 727,457 & 2,955,922\\ 
&& English & 309,352,280 & 52.00 & 0.00 & Latin & 1611 & 762,180 & 2,800,404\\ 
&& German & 95,392,978 & 52.00 & 10.00 & Latin & 1545 & 698,033 & 2,923,630 \\ 
&& Norwegian & 4,320,000 & 61.00 & 8.00 & Latin & 1904 & 727,315 & 2,819,120\\
&&  Swedish & 8,789,835 & 60.00	& 15.00 & Latin & 1917 & 733,503 & 2,978,788 \\ \cline{2-10}
& Greek & Greek (Modern) & 12,258,540 & 39.00	&22.00& Greek & 1840 & 706,871 & 3,054,402\\ \cline{2-10}
& \multirow{7}{2em}{Indic} & \textbf{Bengali} & 70,561,000 & 24.00 & 90.00 & Bengali & - & 603,911 & 1,732,146 \\
&& Gujarati & 46,106,136 & 23.00 & 72.00 & Gujarati & 1823 & 199,287 & 896,718 \\
&& Hindi & 180,764,791 & 25.00 & 77.00& Devanagari & 1818 & 794,341 & 2,870,484 \\
&& Marathi & 68,049,787 & 19.00	& 76.00 & Devanagari & 1821 & 641,455 & 3,389,594 \\
&& Nepali & 17,209,255 & 28.00	& 85.00 & Devanagari & 1914 & 670,706 & 3,108,308 \\ 
&& \textbf{Sindhi} & 21,362,000 &26.00 & 69.00 & Arabic & 2022 & 576,350 & 2,427,864 \\
&& \textbf{Sinhala} & 13,220,256 & 7.00	& 80.50 & Sinhala & - & 457,448 & 2,963,558 \\
\cline{2-10} 

\hline
\end{longtable}
\end{sidewaystable}    

\newpage
\begin{sidewaystable}  
\begin{longtable}{ |c|c|c|c|c|c|c|c|c|c| } 
\hline
\rowcolor{lightgray} Language family & Genus & Language & Speakers &Latitude & Longitude & Script & Year & Words & Phonemes\\
\hline
\multirow{18}{6em}{Indo-European} & \multirow{2}{3em}{Iranian}&Persian & 24,316,121 & 32.00 & 54.00& Arabic & 1838 & 672,783 & 3,064,961  \\
&& \textbf{Tajik}\textsuperscript{\textdagger} & 4,380,212 & 38.67 & 70.00 & Cyrillic & 1992 & 664,267 & 3,092,578 \\ \cline{2-10} 
& \multirow{6}{4em}{Romance} &  \textbf{Catalan} & 6,667,328 & 41.75 & 2.00 & Latin & - & 617,575 & 2,431,494\\
&&French & 64,858,311 & 48.00 & 2.00& Latin & 1776 & 733,500 & 2,196,754\\
&&  Italian & 61,489,984 & 43.00 & 12.00& Latin & 1649 & 667,791 & 2,977,510\\
&& Portuguese & 177,457,180 &39.00	& -8.00& Latin & 1751 & 709,214 & 2,955,437\\
&& Romanian & 23,498,367 & 46.00 & 25.00& Latin & 1928 & 703,169 & 1,372,925\\
&&  Spanish & 322,299,171 & 40.00 & -4.00& Latin & 1569 & 719,118 & 3,026,226 \\
\cline{2-10}
& \multirow{10}{2em}{Slavic} & \textbf{Belorussian} & 9,081,102 &54.00	&28.00& Cyrillic & - & 568,227 & 1,903,649\\ 
&& \textbf{Bosnian} & 4,000,000 & 43.00	& 18.00 & Latin & - & 583,203 & 2,675,548\\
&& Bulgarian & 8,954,811 &42.50	& 25.00& Cyrillic & 1864 & 555,959 & 2,941,503 \\
&& Czech & 11,525,089 & 50.00 & 15.00& Latin & 1380 & 576,091 & 2,819,779\\
&& Polish & 42,708,133 & 52.00 & 20.00& Latin & 1975 & 601,693 & 2,877,812 \\
&& Russian & 145,031,551 &56.00	& 38.00 & Cyrillic & 1876 & 567,632 & 2,790,746 \\
&& Serbian & 11,144,758 &44.00 & 19.00& Latin & 1804 & 585,831 & 2,539,211 \\
&& Slovene & 1,984,775 &46.00 & 15.00 & Latin & 1584 & 630,070 & 2,942,416 \\
&& Slovak & 5,011,120 & 49.00 & 20.00 & Latin & 1832 & 606,161 & 2,870,657 \\
&& Ukrainian & 39,441,842 & 49.00 & 33.00 & Cyrillic & 1903 & 132,657 & 620,712\\
\hline
Nakh-Dagestanian & Avar-Andic-Tsezic & \textbf{Avar}\textsuperscript{\textdagger} & 600,959 & 42.50 & 46.50 & Latin & - & 96,114 & 676,715 \\
\hline
Niger-Congo & Bantu & \textbf{Bukusu}\textsuperscript{\textdagger} & 565,000 & 0.75 & 34.67 & Latin & - & 560,629 & 3,158,798 \\
\hline
\multirow{3}{4em}{Uralic} & \multirow{2}{3em}{Finnic} &  Estonian & 1,075,497 & 59.00 & 26.00& Latin & 1739 & 153,645 & 737,343 \\
&&  Finnish & 5,232,728 & 62.00	& 25.00 & Latin & 1776 & 543,872 & 3,183,574\\ \cline{2-10}
& Ugric & Hungarian & 13,611,600 & 47.00 & 20.00 & Latin & 1590 & 589,115 & 2,982,891  \\ \hline
\end{longtable}
\end{sidewaystable}    
\clearpage
\newpage

\section{Alternative distance metrics} \label{app_alt}
\begin{figure}[t]
    \centering
    \includegraphics[width=.85\textwidth,clip]{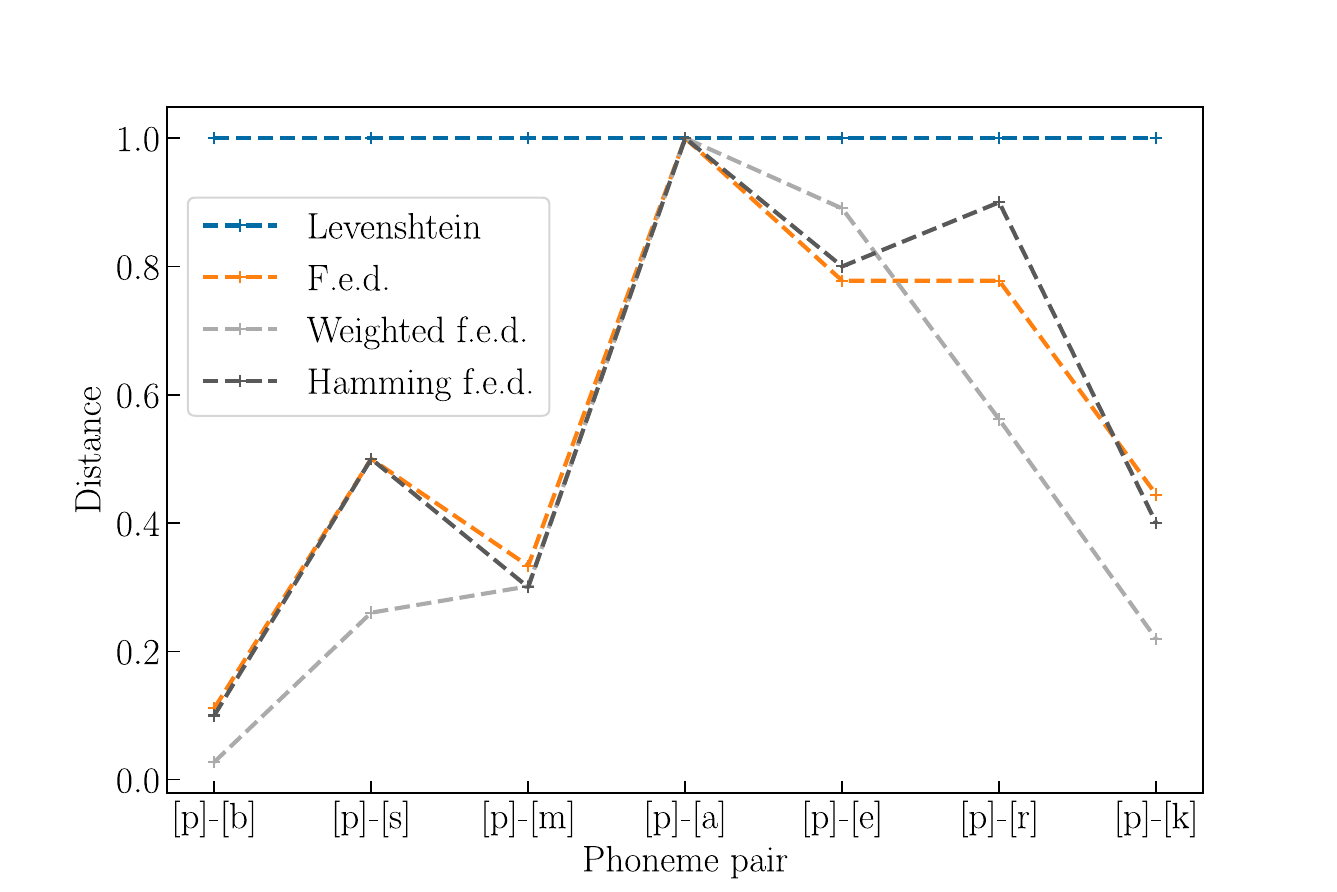}
\caption{Comparison of four different distances between the
feature vector representations of selected phoneme pairs. Each
curve is normalized to its maximum value. F.\ (f.) e.\ d.\
stands for feature edit distance.}
\label{distancecomparison}
\end{figure}

Here we discuss different metrics that allow us to calculate the articulatory-feature distance between our 72-dimensional
3-phone vectors. In particular, we compare in Fig.~\ref{distancecomparison} Levenshtein, Hamming and two feature (weighted and unweighted) edit distances~\cite{mortensen_panphon_2016}
for seven characteristic phoneme pairs.
The feature edit distances attribute a lower cost to edits from specified to unspecified and vice versa. This
cost is uniform for the unweighted distance but feature-dependent for the weighted distance. The Hamming distance, lastly, attributes the same cost to all edits, and higher cost for substitutions. We notice in Fig.~\ref{distancecomparison}
that the Levenshtein distance is uniform over our sample, which makes it uninteresting for
the purpose of taking into account phonological similarities. Both the unweighted and Hamming feature edit distances exhibit a similar behavior, largely consistent with, yet different from, the one shown by the weighted distance. Therefore, we select the unweighted feature distance for $d$ in Eq.~\eqref{eq_W} since
this metric offers comparable performance while being computationally less demanding.

\bibliography{biblio}

\end{document}